\documentclass[letterpaper, 10pt, conference]{ieeeconf}
\IEEEoverridecommandlockouts 

\usepackage{times}
\usepackage{hyperref}
\usepackage{graphicx}
\usepackage{amsfonts}
\usepackage{amsmath}
\usepackage{amssymb}
\usepackage{array}
\usepackage{flafter}
\usepackage{cite}
\usepackage{subfigure}
\usepackage{verbatim}
\usepackage{multirow}
\usepackage{multicol}
\usepackage{cleveref}
\Crefname{equation}{Eq.}{Eqs.}
\Crefname{figure}{Fig.}{Figs.}
\Crefname{tabular}{Tab.}{Tabs.}
\usepackage[usenames,dvipsnames]{color}
\usepackage[ruled,vlined,linesnumbered]{algorithm2e}

\title{\huge Autonomous Exploration Development Environment \\ and the Planning Algorithms
\thanks{All authors are with the Robotics Institute at Carnegie Mellon University, Pittsburgh PA. Emails: {\tt \{ccao1, hongbiaz, fanyang2, yukunx, choset, jeanoh, zhangji\}@cmu.edu}}}
\author{Chao Cao, Hongbiao Zhu, Fan Yang, Yukun Xia, Howie Choset, Jean Oh, and Ji Zhang}

\begin{document}

\maketitle


\begin{abstract}
Autonomous Exploration Development Environment is an open-source repository released to facilitate development of high-level planning algorithms and integration of complete autonomous navigation systems. The repository contains representative simulation environment models, fundamental navigation modules, e.g., local planner, terrain traversability analysis, waypoint following, and visualization tools. Together with two of our high-level planner releases -- TARE planner for exploration and FAR planner for route planning, we detail usage of the three open-source repositories and share experiences in integration of autonomous navigation systems. We use DARPA Subterranean Challenge as a use case where the repositories together form the main navigation system of the CMU-OSU Team. In the end, we discuss a few potential use cases in extended applications.
\end{abstract}

\begin{figure}[t]
	\vspace{-0.1in}
    \centering
    \includegraphics[width=0.8\linewidth]{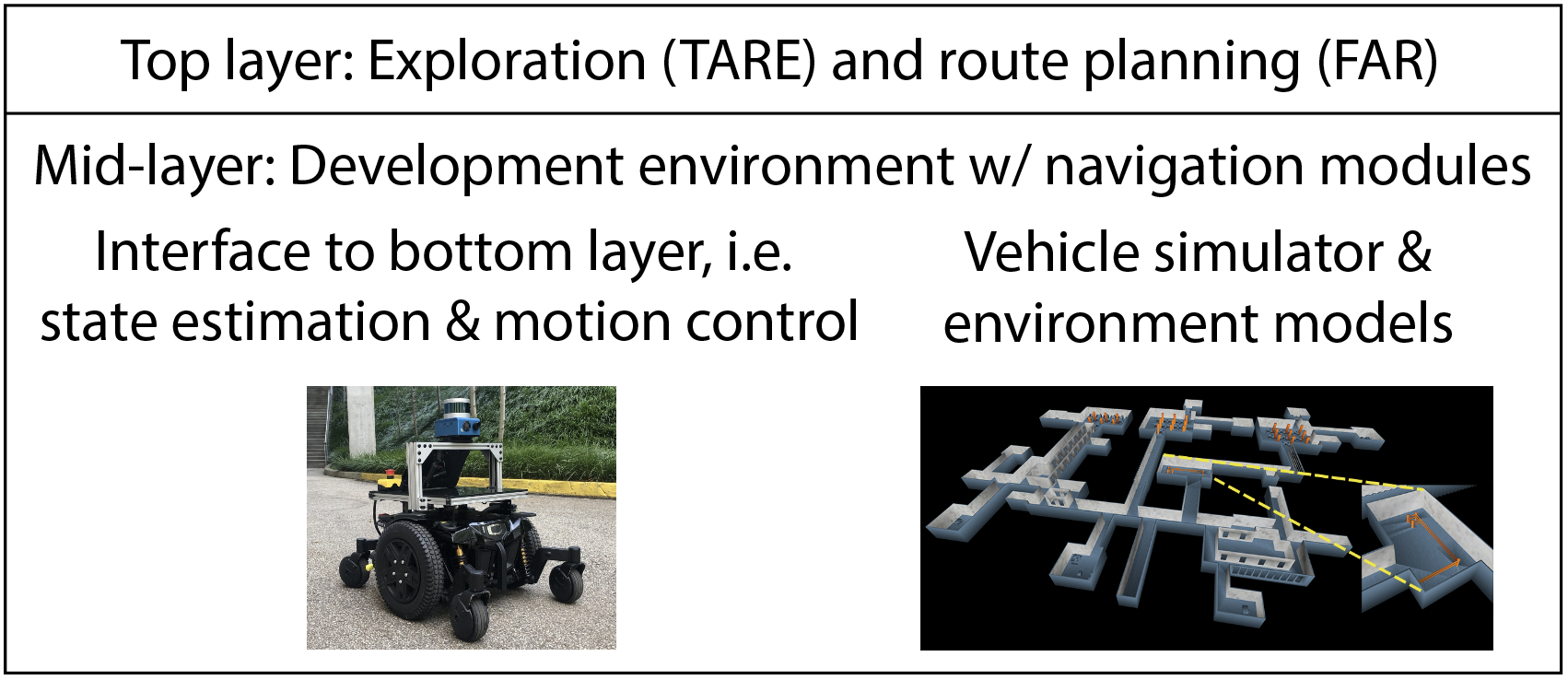}
    \caption{Our repositories form the mid-layer and top layer in a three-tier navigation system. Users can integrate the repositories in real systems using the interface to bottom layer, i.e., state estimation and motion control, or run the repositories in simulation using the environment models.}
	\label{fig:opening}
	\vspace{-0.15in}
\end{figure}

\section{Introduction}\label{sec:intro}

Our Autonomous Exploration Development Environment\footnote{Development Environment: \url{www.cmu-exploration.com}}, exploration planner, namely TARE planner\footnote{TARE Planner: \url{github.com/caochao39/tare_planner}} \cite{cao2021tare}, and route planner, namely FAR planner\footnote{FAR Planner: \url{github.com/MichaelFYang/far_planner}} \cite{yang2022far}, form a full stack of geometry-based algorithms for navigation planning. Combining the three repositories, we provide a generic set of 3D Lidar-based navigation algorithms for ground vehicles. The software stack is beneficial to the research society as a platform for developing and deploying advanced navigation systems, and further meant to support state-of-the-art research in vision-based navigation.


The development environment is made to facilitate development of high-level planning algorithms.
We provide a set of environment models for users to start algorithm development conveniently. The development environment is further compatible with photorealistic house models from Matterport3D \cite{Matterport3D}. We provide a set of fundamental navigation modules, e.g., local planner for collision avoidance, terrain traversability analysis, waypoint following. In a standard three-tier autonomous navigation system, the development environment functions as the mid-layer, interfacing with the state estimation and motion control on the bottom layer, and the high-level planner on the top layer. For providing state estimation to the system, we made available a list of compatible open-source state estimation methods.

Our navigation algorithms are created with favoring real-system deployment kept in mind. We do not rely on any unrealistic information which can be made available only in simulation, e.g. semantic and terrain segmentation ground-truth. In particular, our terrain traversability is computed by a dedicated module that processes range data as in a real system. The basic concept is allowing users to develop high-level planning algorithms and conveniently migrate the repositories to the vehicle computer for deployment.

The repositories of the development environment, TARE planner, and FAR planner together form the main navigation system that the CMU-OSU Team uses to attend the DARPA Subterranean Challenge. In the final competition, the team received a ``Most Sectors Explored Award" by conducting the most complete exploration (26 out of 28 sectors) among all teams. 
This paper uses the system as an example to detail usage of the repositories in autonomous exploration and navigation. We further discuss potential use cases in extended applications in simulated and real systems.

\section{Related Work}\label{sec:related}

Autonomous navigation systems have been studied from multiple angles. The work described in this paper is based on key results in datasets, simulation environments, and navigation systems briefly discussed in this section.


\textit{Datasets}: For outdoor settings, the most well-known dataset is KITTI Vision Benchmark Dataset \cite{geiger2013vision} collected from a self-driving suite in road driving scenarios. The dataset contains sensor data from stereo cameras, 3D Lidar, and GPS/INS ground truth for benchmarking depth reconstruction, odometry estimation, object segmentation, etc. For localization purposes, the long-term localization benchmark \cite{zhang2021reference, sattler2018benchmarking} includes a comprehensive list of datasets such as Aachen Day-Night dataset, Extended CMU Seasons dataset, RobotCar Seasons dataset, and SILDa Weather and Time of Day dataset. For indoor scenes, datasets such as NYU-Depth dataset \cite{silberman2012indoor}, TUM SLAM dataset \cite{sturm2012benchmark}, inLoc dataset \cite{taira2018inloc}, MIT Stata center dataset \cite{fallon2013stata}, and KTH-INDOL dataset \cite{pronobis2006discriminative, luo2007incremental} are available. Datasets are useful in developing and benchmarking perception and planning algorithms providing users real sensor readings especially for those who don't have access to integrated sensor suites.

\textit{Simulation environments}: Carla\cite{dosovitskiy2017carla} and AirSim \cite{shah2018airsim} are two representative simulation environments for autonomous driving and flying. These simulators support various conditions such as lighting and weather changes, moving objects such as pedestrians, and incident scenes. For indoor navigation, iGibson \cite{shen2020igibson, xia2020interactive}, Sapien \cite{xiang2020sapien}, AI2Thor \cite{kolve2017ai2}, Virtual Home \cite{puig2018virtualhome}, ThreeDWorld \cite{gan2020threedworld}, MINOS \cite{savva2017minos}, House3D \cite{wu2018building} and CHALET \cite{yan2018chalet} use synthetic scenes, while reconstructed scenes are also available in iGibson, AI Habitat \cite{savva2019habitat, szot2021habitat} and MINOS \cite{savva2017minos}. Compared to datasets, simulation environments have the advantage of providing access to ground truth data, e.g. vehicle pose and semantic segmentation to simplify the algorithm development and allowing full navigation system tests in closed control loop.


\textit{Navigation systems}:
While countless algorithms have been developed for autonomous navigation, we only list the system-level efforts providing a comprehensive set of open-source perception and planning modules to be integrated as a navigation system. ROS (Robot Operating System) Navigation Stack \cite{rosnavstack} is an entry-level mobile robot navigation solution. The stack consists of packages including state estimation and planning modules to conduct navigation in 2D known/unknown environments.
The open-source repositories from ETH Autonomous System Lab \cite{ethlab} and HKUST Aerial Robotics Group \cite{hkustlab} provide packages ranging from sensor drivers to state estimation \cite{leutenegger2015keyframe} \cite{qin2018online}, planning \cite{oleynikova2017voxblox,zhou2019robust}, and exploration \cite{bircher2016receding, zhou2021fuel} methods. Users can use these packages to integrate aerial navigation systems.

\textit{Our navigation system}: Our system follows the three-tier architecture. Our development environment servers as the mid-layer in the system providing generic interfaces to state estimation and motion control modules as well as high-level planning algorithms. The development environment provides a set of fundamental navigation modules such as local planner for collision avoidance, terrain traversability analysis, and waypoint following, and supports common differential-drive platforms. The purpose is using the development environment to facilitate users to develop high-level planning algorithms and then port the entire repositories to the vehicle computer for deployment. Our environment models are both indoor and outdoor containing complex topology to stress exploration and navigation algorithms. In addition to our environment models, we support the photorealistic house models from Matterport3D \cite{Matterport3D} and provide interface to AI Habitat \cite{savva2019habitat, szot2021habitat}. Both are widely used by the robotics and computation vision societies. We aim at making our system useful to support research in those societies.

\begin{figure}[t]
	\centering
	\subfigure[Campus]{\includegraphics[width=0.48\linewidth]{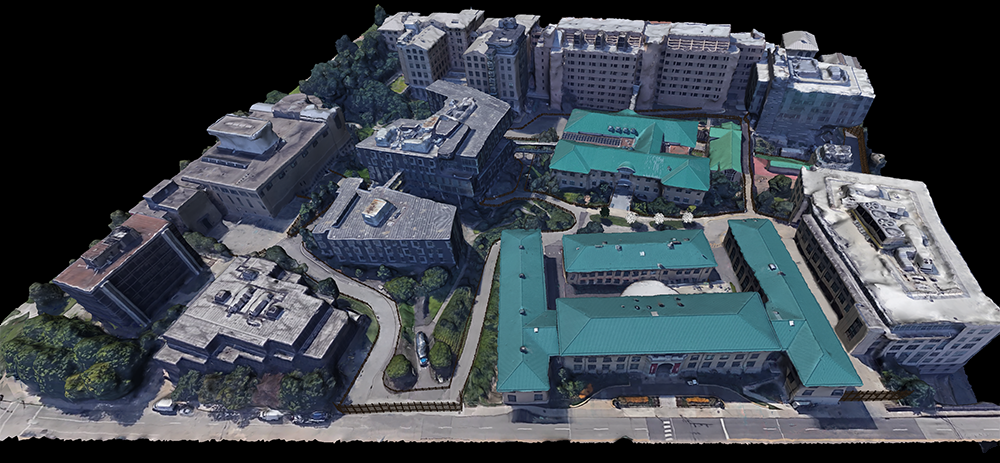}} \hspace{0.01in}
	\subfigure[Indoor]{\includegraphics[width=0.48\linewidth]{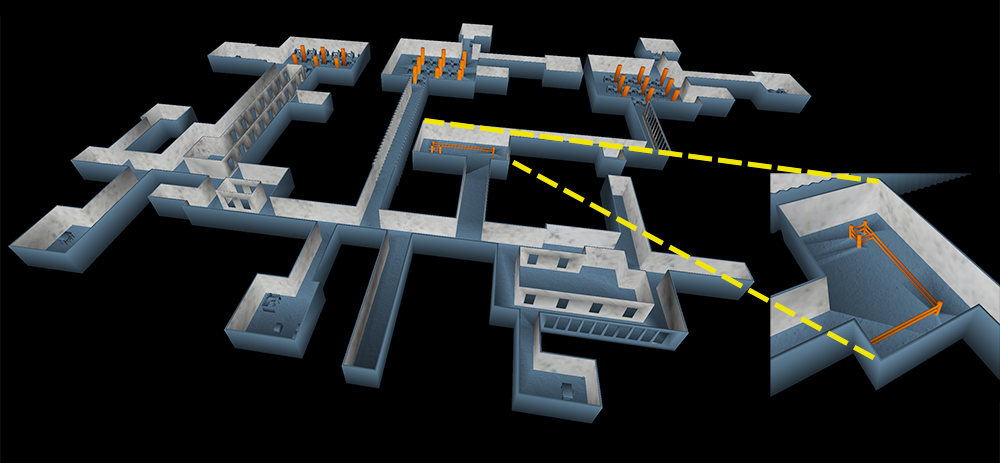}} \\\vspace{-0.05in}
	\subfigure[Garage]{\includegraphics[width=0.48\linewidth]{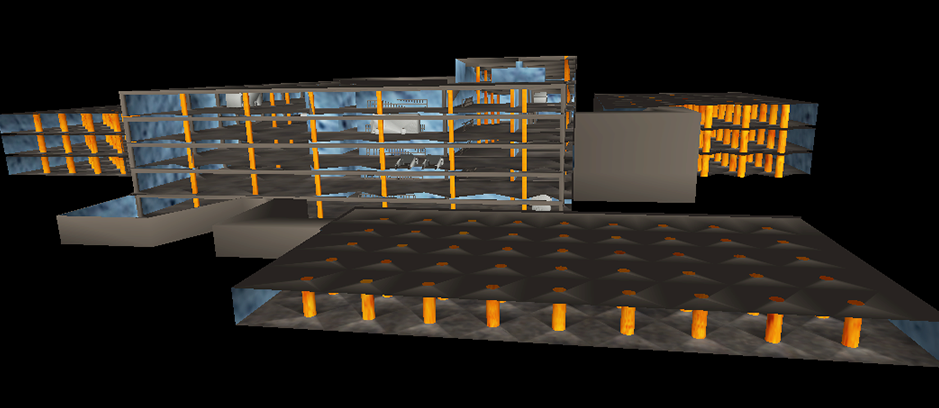}} \hspace{0.01in}
	\subfigure[Tunnel]{\includegraphics[width=0.48\linewidth]{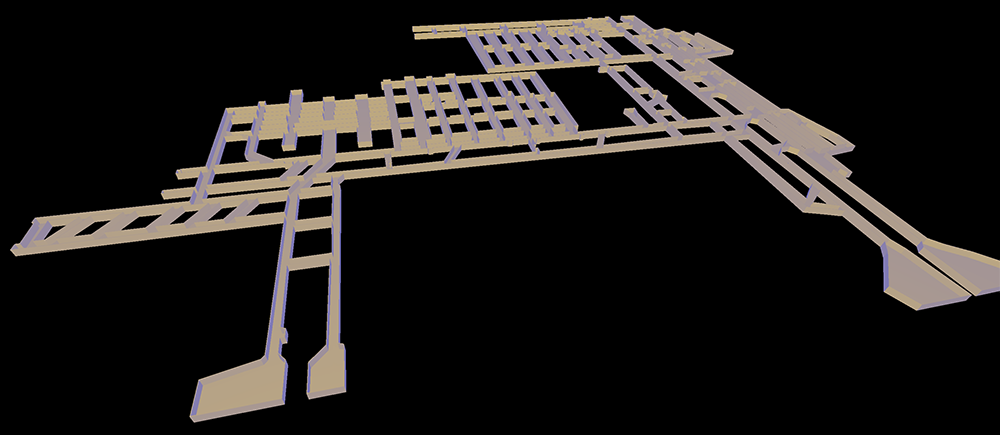}} \\\vspace{-0.05in}
	\subfigure[Forest]{\includegraphics[width=0.48\linewidth]{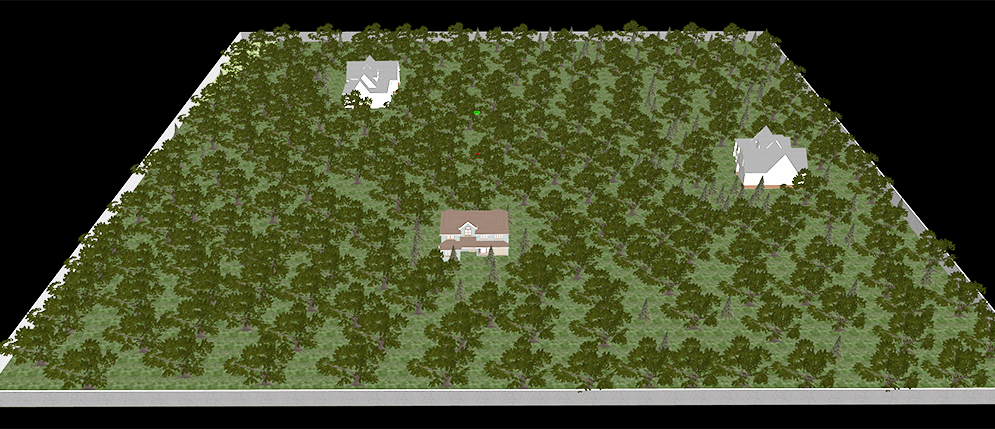}} \\\vspace{-0.05in}
	\caption{Five environment models. The characteristics of the environment models are listed in Table \ref{tab:environment}.}
	\label{fig:simulation_environment}
 	\vspace{-0.15in}
\end{figure}

\section{Development Environment}\label{sec:development_env}

The development environment functions as a platform for developing and benchmarking high-level planning algorithms for ground vehicle navigation. In the development environment, we provide five environment models, fundamental navigation modules, and visualization and debugging tools.

\subsection{Environment Models}

The environment models resemble real-world settings where robotic systems are commonly deployed. Each of the environment models is distinctive with unique features and challenges. Fig.~\ref{fig:simulation_environment} gives an overview of the environment models and Table \ref{tab:environment} summarizes their characteristics.

\begin{table}[b]
\vspace{-0.15in}
\renewcommand{\arraystretch}{1.15} 
\centering{\scriptsize
\caption{Environment model characteristics} \label{tab:environment}
\vspace{-0.05in}
\begin{tabular}{p{0.7cm}p{0.4cm}p{0.9cm}p{0.7cm}p{0.9cm}p{0.9cm}p{0.9cm}}
& Large & Convoluted & Multi & Undulating & 
Cluttered &
Thin \\
& Scale &
 &
Storage &
Terrain &
Obstacles &
Structure \\
\hline 
Campus & \multicolumn{1}{|c}{X} & X & & X & & \\
Indoor & \multicolumn{1}{|c}{} & X & & & X & X \\
Garage & \multicolumn{1}{|c}{} & & X & X & & \\
Tunnel & \multicolumn{1}{|c}{X} & X &  &  & & \\
Forest & \multicolumn{1}{|c}{} &  &  &  & X & \\
\end{tabular}}
\end{table}


\begin{itemize}
    \item \textit{Campus} (340m $\times$ 340m): A large-scale environment as part of the Carnegie Mellon University campus, containing undulating terrains and convoluted layout.
    \item \textit{Indoor} (130m $\times$ 100m): Consists of long and narrow corridors connected with lobby areas. Obstacles such as tables and columns are present.
    \item \textit{Garage} (140m $\times$ 130m, 5 floors): An environment with multiple floors and sloping terrains to test autonomous navigation in a 3D environment.
    \item \textit{Tunnel} (330m $\times$ 250m): A large-scale environment containing tunnels that form a network, provided by the Autonomous Robots Lab at University of Nevada, Reno.
    \item \textit{Forest} (150m $\times$ 150m): Containing mostly trees and a couple of houses in a cluttered setting.
\end{itemize}

\begin{figure}[t]
    \vspace{0.05in}
	\centering
	\subfigure[]{\includegraphics[width=0.75\linewidth]{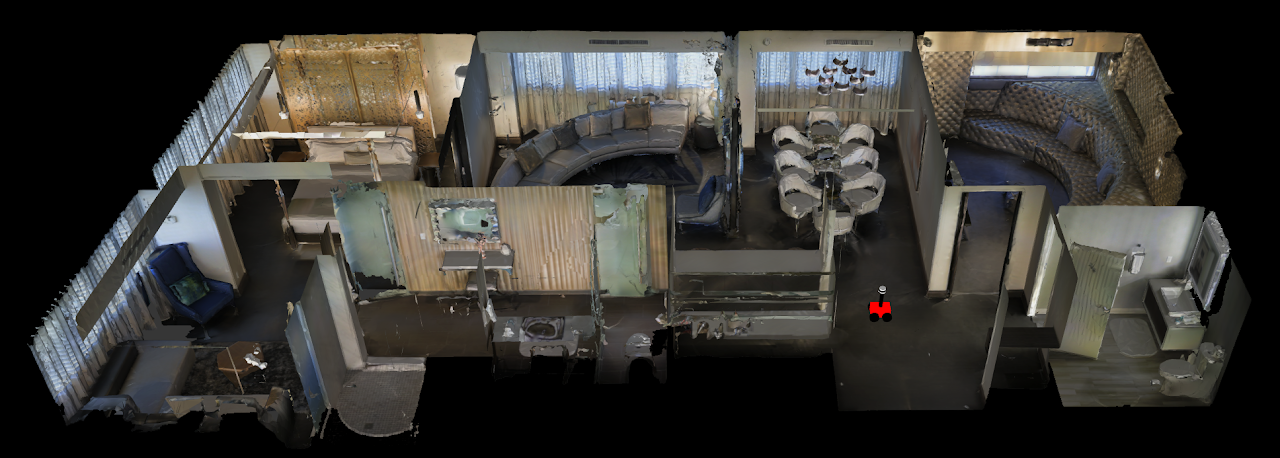}} \\\vspace{-0.05in}
	\subfigure[]{\includegraphics[width=0.3\linewidth]{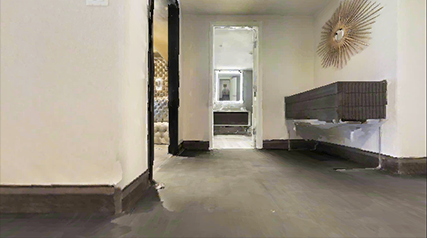}} \hspace{0.005in}
	\subfigure[]{\includegraphics[width=0.3\linewidth]{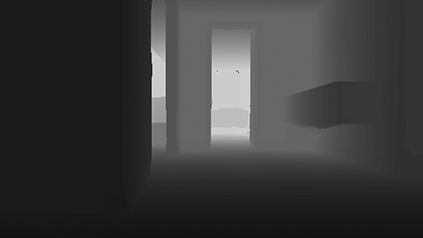}} \hspace{0.005in}
	\subfigure[]{\includegraphics[width=0.3\linewidth]{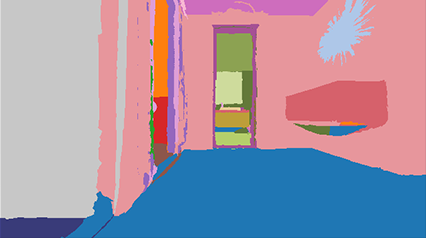}} \\\vspace{-0.05in}
    \caption{(a) A Matterport3D house model and (b) corresponding RGB, depth, and semantic images rendered by AI Habitat.}
	\label{fig:matterport}
 	\vspace{-0.15in}
\end{figure}

Our system also supports the photorealistic house models from Matterport3D \cite{Matterport3D}. Users are provided with scan data and RGB, depth, and semantic images rendered by AI Habitat \cite{savva2019habitat, szot2021habitat}. Users have the option of running AI Habitat side by side with our system or in post-processing. Detailed instructions on configuring our system to use Matterport3D house models and AI Habitat is available on our website.


\subsection{Local Planner}

\begin{figure}[b]
 	\vspace{-0.1in}
	\centering
	\subfigure[]{\includegraphics[width=0.215\textwidth]{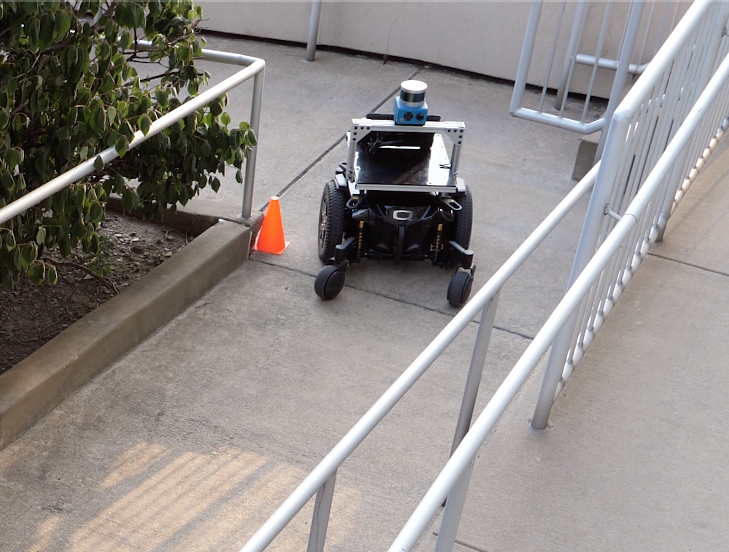}} \hspace{0.05in}
	\subfigure[]{\includegraphics[width=0.20\textwidth]{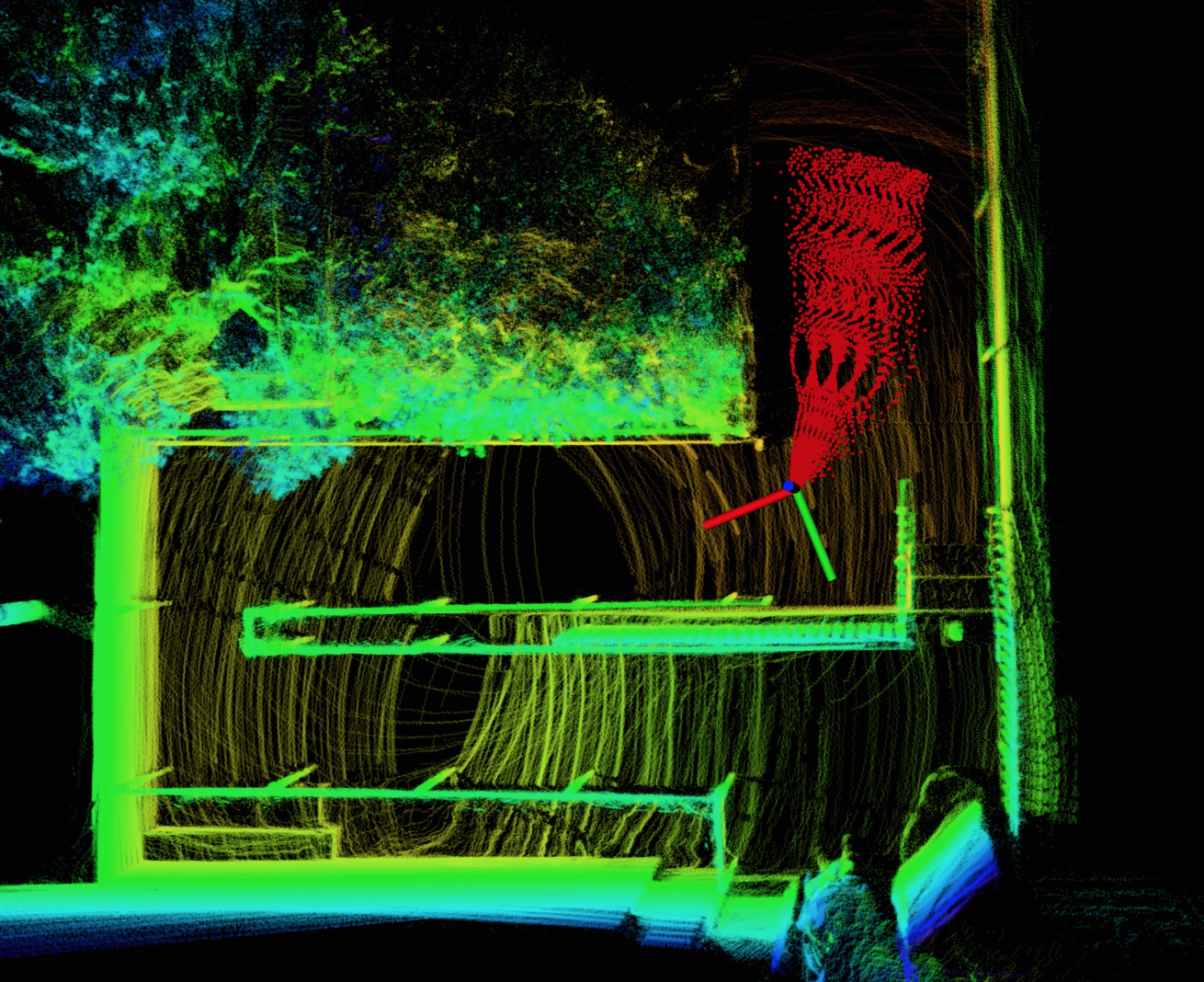}} \\\vspace{-0.05in}
    \caption{Example motion primitives. The vehicle in (a) and the coordinate frame in (b) share the same location. The red paths in (b) indicate collision-free motion primitives.}
	\label{fig:local_planner}
\end{figure}

\begin{figure}[t]
	\centering
	\subfigure[]{\includegraphics[width=0.21\textwidth]{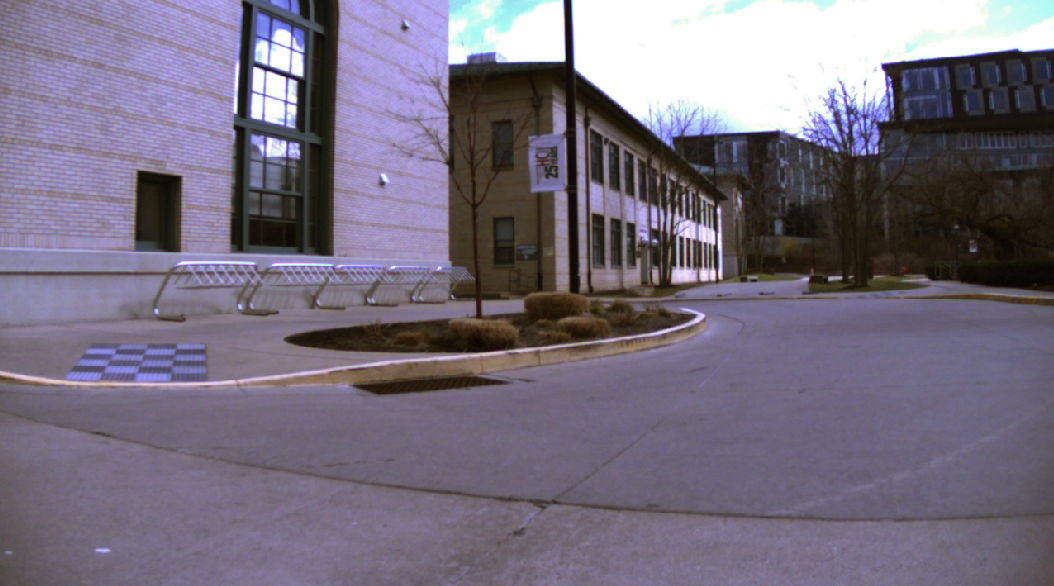}} \hspace{0.05in}
	\subfigure[]{\includegraphics[width=0.21\textwidth]{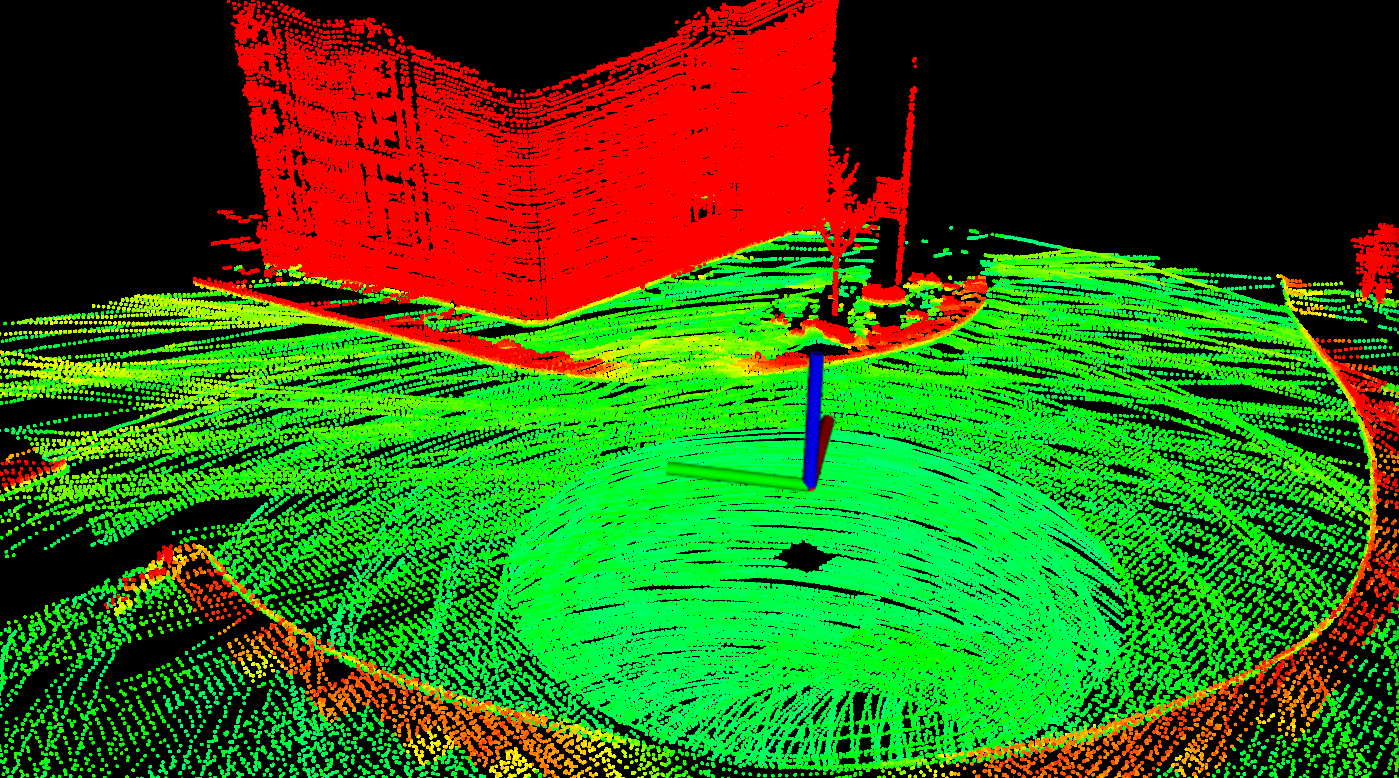}} \\\vspace{-0.05in}
    \caption{Example terrain map. The image in (a) is taken from the location of the coordinate frame in (b). The green points in (b) are traversable and the red points are non-traversable.}
	\label{fig:terrain_map}
 	\vspace{-0.15in}
\end{figure}



The local planner\cite{zhang2020avoidance} warrants safety in reaching waypoints that are sent by high-level planners. It computes and follows collision-free paths that lead to the waypoint. The module pre-computes a motion primitive library and associates the motion primitives to 3D locations in the vicinity of the vehicle. The motion primitives are modeled as Monte Carlo samples and organized in groups. In real-time, when a location is occupied by obstacles, the module can determine motion primitives that are collided with the obstacle within milliseconds. The module then selects the group of motion primitives with the maximum likelihood toward the waypoint. In Fig. \ref{fig:local_planner}, the red paths represent the collision-free motion primitives. For ground vehicles, the traversability of the vehicle is determined by the terrain characteristics. The local planner takes in the terrain map from the terrain analysis module, detailed in the next section. The module also has interface to take in additional range data for collision avoidance as an extension option.


\subsection{Terrain Traversability Analysis}

The terrain analysis module exams the traversability of the local terrain surrounding the vehicle. The module builds a cost map where each point on the map is associated with a traversal cost. The cost is determined by the local smoothness of the terrain. We use a voxel grid to represent the environment and analyze the distributions of data points in adjacent voxels to estimate the ground height. The points are associated with higher traversal costs if they are further apart from the ground. Fig. \ref{fig:terrain_map} gives an example terrain map covering a 40m x 40m area with the vehicle in the center. The green points are traversable and the red points are non-traversable. In addition, the terrain analysis module can handle negative obstacles that often result in empty areas with no data points on the terrain map. When negative obstacle handling is turned on, the module treats those areas as non-traversable.


\subsection{Visualization and Debugging Tools}

To facilitate algorithm development, we provide a set of tools to visualize the algorithm performance. The visualization tools display the overall map, explored areas, and vehicle trajectory. Metrics such as explored volume, traveling distance, and algorithm runtime are plotted and logged to files. Further, the system supports using a joystick controller to interfere with the navigation, switching among multiple operation modes to ease the process of system debugging. Detailed information is available on the project website. 


\begin{figure}[t!]
 	\vspace{0.05in}
    \centering
    \includegraphics[width=0.75\linewidth]{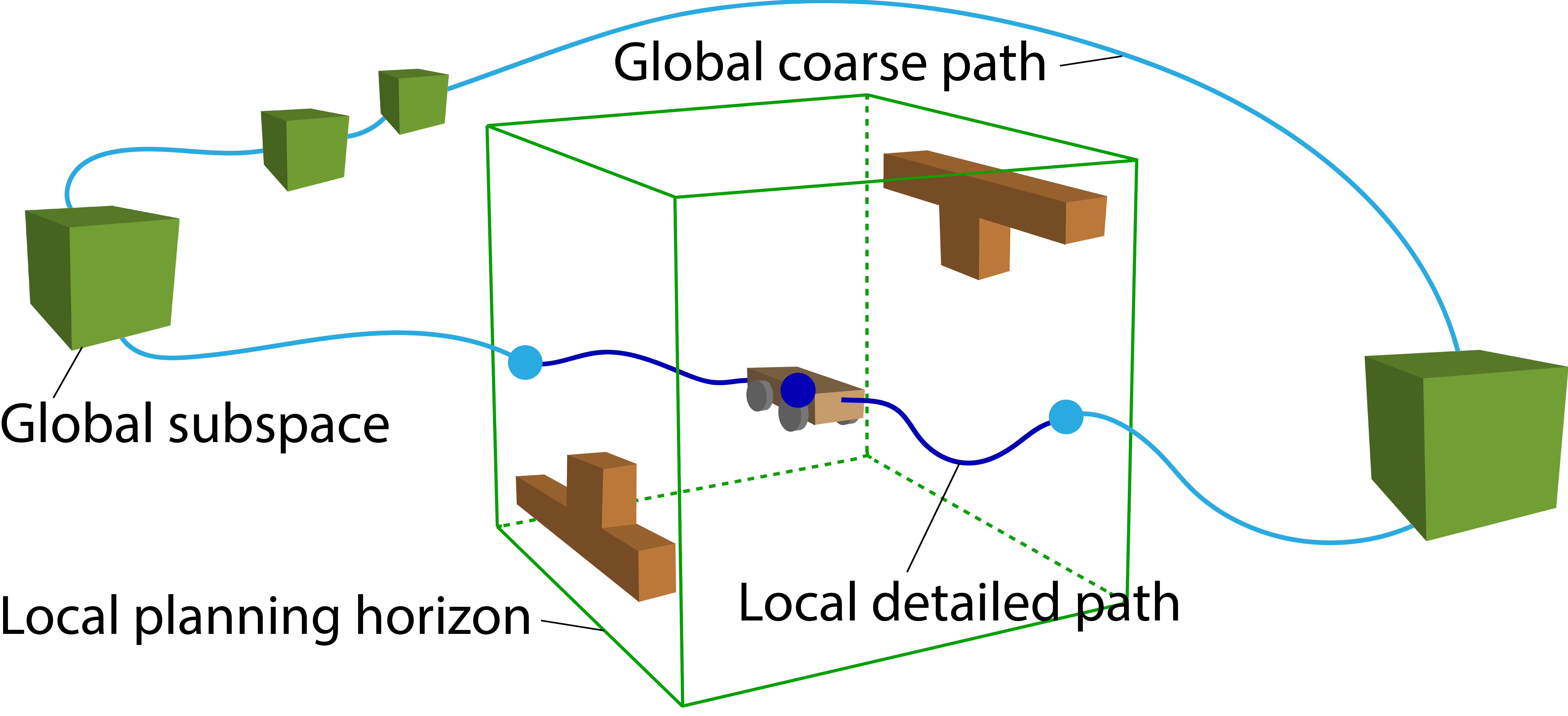}
    \vspace{-0.05in}
    \caption{Illustration of TARE planner's framework. In areas surrounding the vehicle (green box), data is densely maintained and a local detailed path is computed  (dark-blue curve). At the global scale, data is sparsely maintained in distant subspaces (solid green cubes) and a global coarse path is computed (light-blue curve). The local path and global paths are connected to form the exploration path.}
	\label{fig:tare}
	\vspace{-0.15in}
\end{figure}

\begin{figure}[b]
	\vspace{-0.1in}
    \centering
    \includegraphics[width=0.7\linewidth]{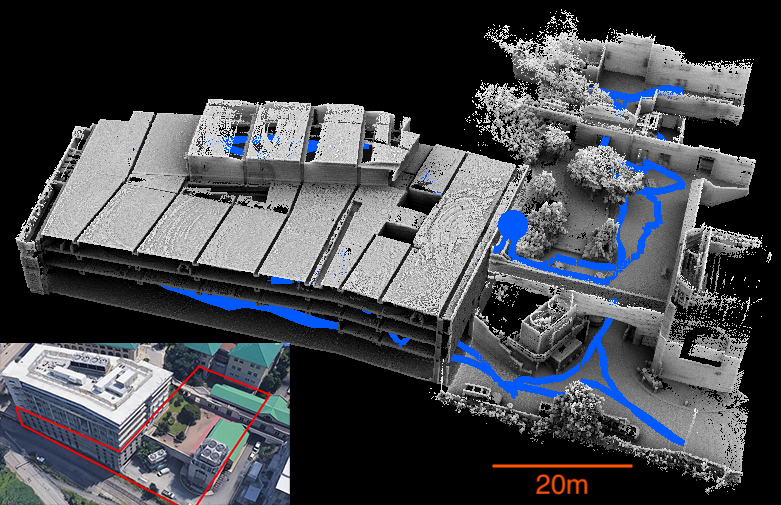}
    \vspace{-0.05in}
    \caption{Resulting map and vehicle trajectory from TARE planner exploring a multi-storage garage and a connected patio. The blue dot is the start point. The red polygon in the bottom-left image shows the location of the site.}
	\label{fig:cic}
\end{figure}

\section{High-level Planners}\label{sec:high_level_planner}

We discuss two high-level planners, TARE planner for exploration and FAR planner for route planning. The high-level planners take in the state estimation output and generate task-specific waypoints, which are executed by the local planner in closed-control-loop navigation. Typically, the high-level planners re-plan at a lower frequency and provide long-distance routes, while the low-level navigation modules react instantaneously to follow the route and avoid obstacles.

\subsection{TARE Planner for Exploration}

TARE planner is a hierarchical framework that utilizes a two-layered representation of the environment to plan the exploration path in a multi-resolution manner. As illustrated in Fig \ref{fig:tare}, the planner uses low-resolution information to plan a coarse path at the global level. In the local area surrounding the vehicle, the planner plans a detailed path using high-resolution information. The method optimizes the overall path by solving a traveling salesman problem at each level. Compared to existing methods relying on greedy strategies, the planner can effectively adapt to the structural environment and produce an approximately optimal exploration path that avoids redundant revisiting.

TARE planner is evaluated in several large and complex environments in both simulation and real world. It is compared to state-of-the-art methods, namely, NBVP\cite{bircher2016receding}, GBP\cite{dang2020graph}, and MBP\cite{dharmadhikari2020motion}. Results indicate that TARE planner produces significantly higher efficiency in exploration and computation. Fig. \ref{fig:cic} shows a representative result where the ground vehicle in Fig.~\ref{fig:local_planner}(a) is used to explore a four-storage garage and a connected patio. The vehicle starts from the entrance of the garage on the top floor and explores the whole environment before reporting completion after 1839m of travel in 1907s. Fig. \ref{fig:cic} shows the resulting map and vehicle trajectory. Due to space limitations, we omit the experiment details. Please refer to our paper \cite{cao2021tare} and website for detailed evaluations of the exploration performance.



\begin{figure}[t]
	\centering
	\subfigure[]{\includegraphics[width=0.7\linewidth]{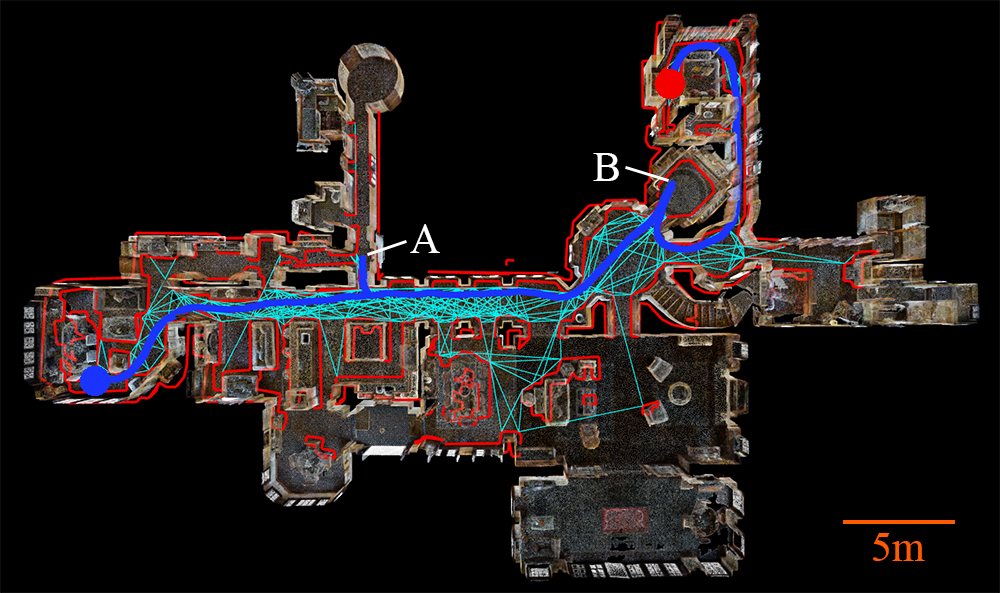}} \\\vspace{-0.05in}
	\subfigure[]{\includegraphics[width=0.7\linewidth]{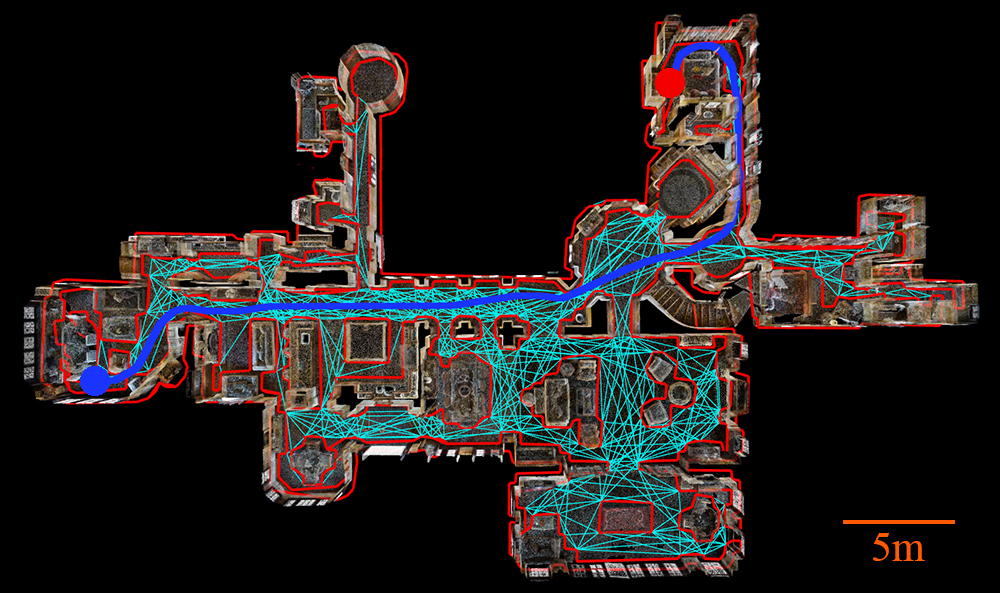}} \\\vspace{-0.05in}
    \caption{Result of FAR planner in (a) unknown and (b) known environments using a Matterport3D house model. The blue curve is the vehicle trajectory staring at the blue dot and ending at the red dot. In (a), the planner attempts to guide the vehicle to the goal by registering obstacles in the environment (red polygons) and building a visibility graph (cyan lines) along with the navigation. At A and B, it re-plans after discovering a better route. In (b), the planner uses the visibility graph from a prior map to plan the route.}
	\label{fig:far_result}
 	\vspace{-0.15in}
\end{figure}

\subsection{FAR Planner for Route Planning}

FAR planner is a visibility graph-based planner that dynamically builds and maintains a reduced visibility graph along with the navigation. The planner can handle both known and unknown environments. In a known environment, it uses a prior map to plan the route. In an unknown environment, however, it attempts multiple ways to guide the vehicle to the goal and picks up the environment layout during the navigation. FAR planner models obstacles in the environment as polygons. It extracts edge points around the obstacles and converts the edge points into a set of polygons. The polygons are then merged over sensor data frames, and from which, the visibility graph is developed.

FAR planner is evaluated in large and convoluted environments. It is compared to RRT* \cite{karaman2011sampling}, RRT-connect \cite{kuffner2000rrt}, A* \cite{hart1968formal}, and D* Lite \cite{koenig2002d} planners and demonstrates its strength in fast re-planing over long distances. The planner uses $\sim$15\% of a single CPU thread on an i7 computer for expanding the visibility graph and a second CPU thread for path search. A path is found within 0.3ms in all of our experiments. Fig.~\ref{fig:far_result} shows a representative result using a Matterport3D house model as the environment. In Fig.~\ref{fig:far_result}(a), the planner does not use any prior information of the environment and treats the environment as unknown. The planner attempts to guide the vehicle to the goal by dynamically registering obstacles in the environment (red polygons) and building a visibility graph (cyan lines) during the navigation. In Fig.~\ref{fig:far_result}(b) the planner is given a prior map. It navigates based on the visibility graph developed from the prior map and plans the route in a known environment.

\begin{figure}[t]
	\centering
	\subfigure[]{\includegraphics[width=0.2\textwidth]{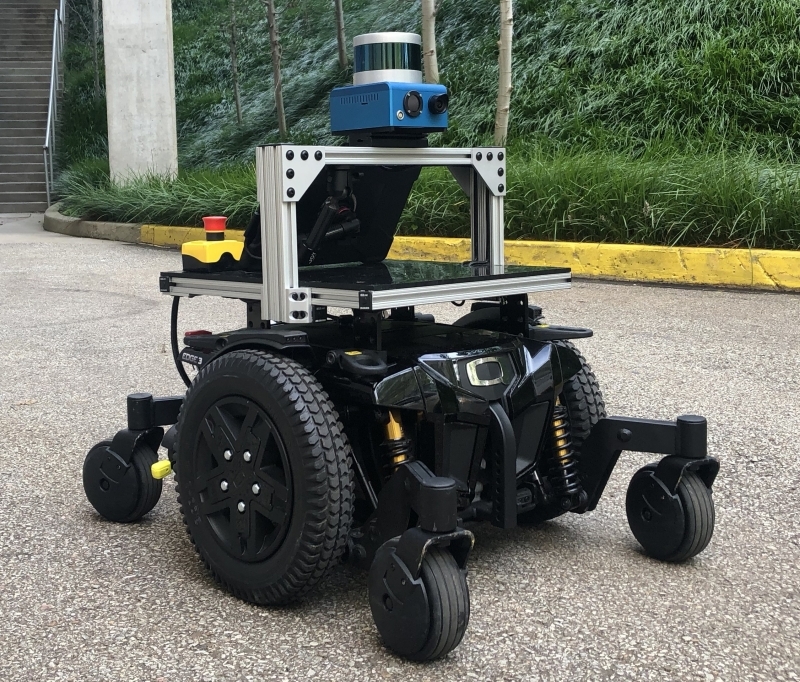}} \hspace{0.05in}
	\subfigure[]{\includegraphics[width=0.23\textwidth]{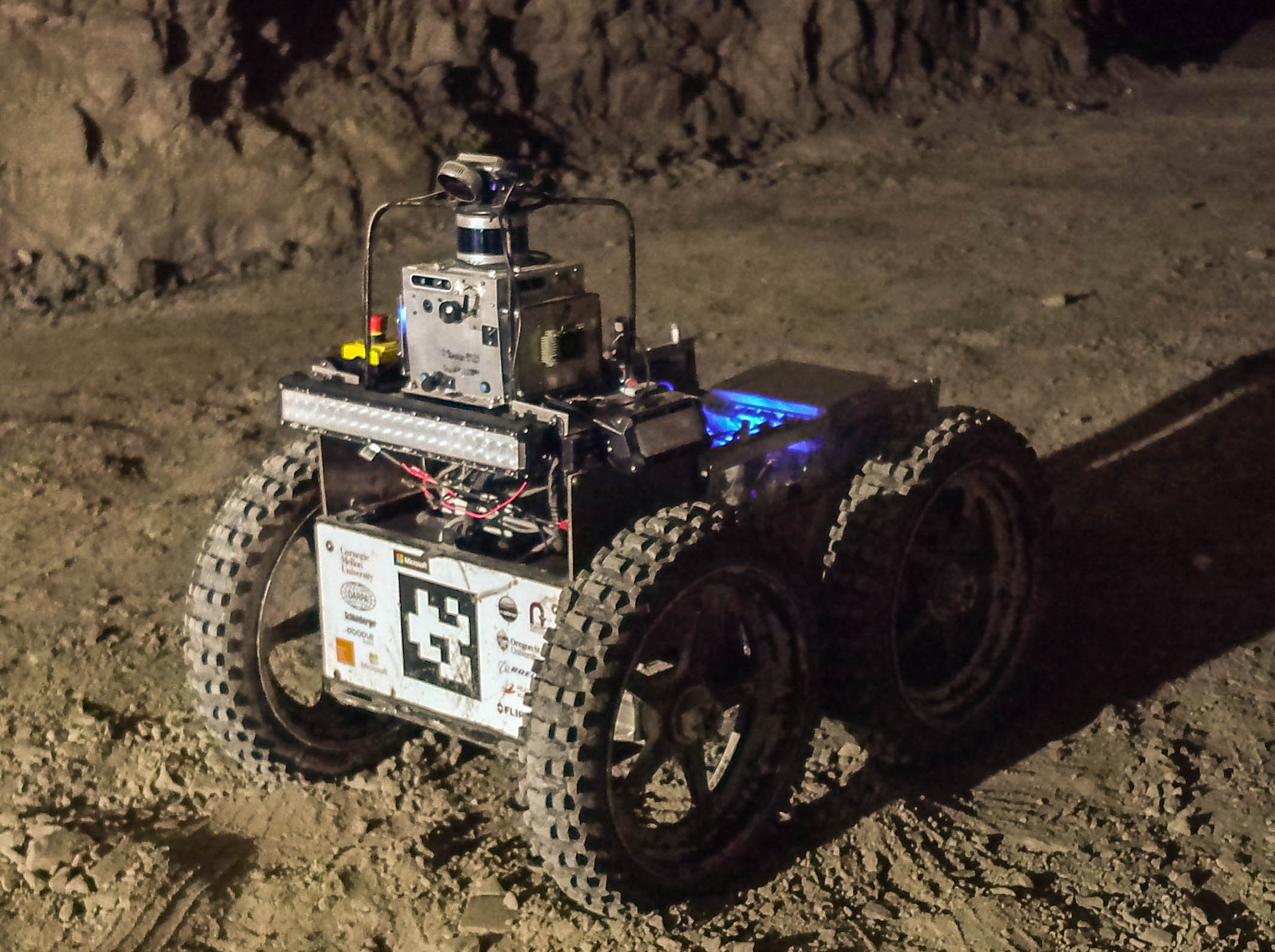}} \\\vspace{-0.05in}
    \caption{(a) Wheelchair-based platform. (b) Ground vehicle used by the CMU-OSU Team in DARPA Subterranean Challenge.}
	\label{fig:robots}
 	\vspace{-0.15in}
\end{figure}

\section{Best Practices}\label{sec:detail}


\textit{Safety margin}: The local planner uses a planning horizon at the distance between the vehicle and waypoint. This ensures that the vehicle can stop at waypoints relatively close to obstacles - collision check does not consider obstacles further than the waypoint. However, if the vehicle is not expected to stop at the waypoint, it is preferable for the high-level planner to keep the waypoint a distance away ($\geq$ 3.75m as the default planning horizon) from the vehicle. If the waypoint is closer, users can project the waypoint further away and keep the waypoint in the same direction w.r.t. the vehicle to fully use the safety margin. On the other hand, if the vehicle needs to navigate through narrow openings, reducing the distance helps the local planner find collision-free motion primitives through the openings.

\textit{Sharp turns}: Typically, a high-level planner selects the waypoint along the path that is a distance, namely look-ahead distance, ahead of the vehicle and sends the waypoint to the local planner (possibly after projecting the waypoint further away from the vehicle as discussed above). When handling sharp turns ($\geq$ 90 deg), the look-ahead distance needs to be properly set or the waypoint may jump to the back of the vehicle, causing the vehicle to osculate back-and-forth. We recommend to select the waypoint on the starting segment of the path that is in line-of-sight from the vehicle.

\textit{Dynamic obstacles}: The terrain analysis module eliminates dynamic obstacles from the terrain map by ray-tracing after the dynamic obstacles move away. This is implemented in the vicinity of the vehicle ($\leq$ 5m to the vehicle) due to the fact that range data becomes sparse further away and the trade-off between eliminating dynamic obstacles and thin structures is hard to set. Depending on the use case, users are advised to perform task-specific dynamical obstacle handling. Both TARE and FAR planners contain such a step.

\begin{figure}[b]
 	\vspace{-0.1in}
    \centering
    \includegraphics[width=0.7\linewidth]{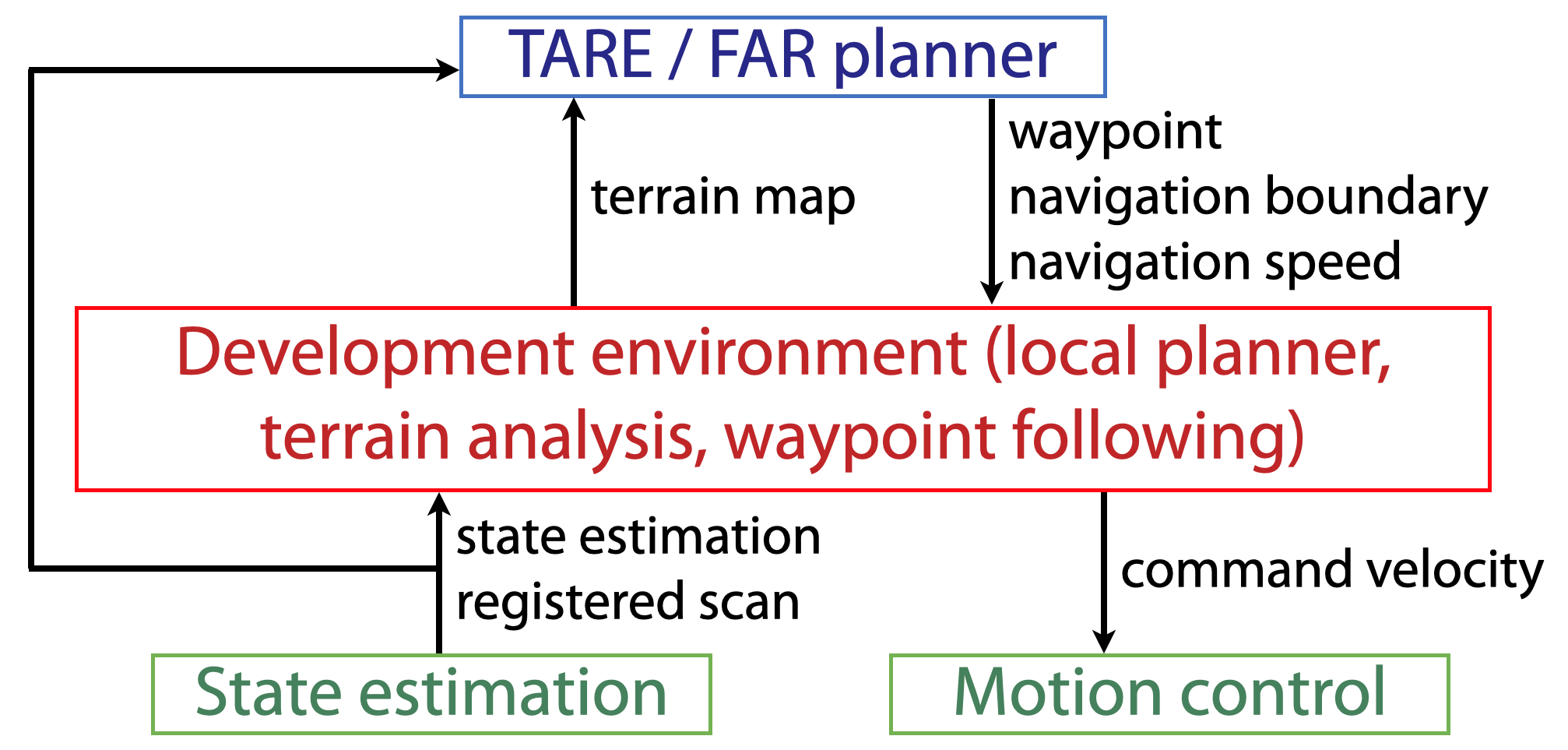}
    \vspace{-0.05in}
    \caption{Basic system diagram for wheelchair-based platform.}
	\label{fig:system_diagram}
\end{figure}


\section{System Integration}\label{sec:system}


\subsection{Basic System}\label{sec:system_wheelchair}

Our basic system uses a wheelchair-based platform shown in Fig. \ref{fig:robots}(a). The vehicle is equipped with a Velodyne Puck Lidar, a camera at 640 $\times$ 360 resolution, and a MEMS-based IMU. A 4.1GHz i7 computer handles processing onboard. Fig.~\ref{fig:system_diagram} shows the basic system diagram. The development environment functions as the mid-layer where it takes in the output from state estimation module and sends command velocity to the motion control module on the bottom layer. It interfaces with the high-level planner (TARE or FAR planner) at the top layer by providing terrain map and taking in waypoint, navigation boundary, and speed. Such a system setup allows users to develop high-level planning algorithms without the necessity of understanding the interfaces among low-level navigation modules.



Our system is compatible with several open-source 3D Lidar-based odometry/SLAM methods as candidates of the state estimation module, including LOAM \cite{zhang2014loam}, A-LOAM \cite{aloam}, LeGO-LOAM \cite{shan2018lego}, LIO-SAM \cite{shan2020lio} and LIO-mapping \cite{ye2019tightly}. Instructions on setting up these methods can be found on our website. Specifically, our system requires the state estimation module to output scan data registered in the world frame. Using registered scans has a few advantages, i.e., it makes processing less sensitive to the time synchronization between scan data and state estimation. Further, multiple registered scans can be stacked together to extract rich geometric information from the environment. However, if scan data associated with the sensor frame is needed, our system also provides it with the corresponding synchronized state estimation. Our system supports generic differential-drive platforms (including skid-steer mechanisms), which is the most commonly used mobile robot kinematic model. Omnidirectional and legged platforms can still use our system but their mobility is not fully exploited, i.e., they move as a differential-drive platform without lateral movement. Our system currently does not support car-like platforms.

\subsection{DARPA Subterranean Challenge System}\label{sec:system_subt}

DARPA Subterranean Challenge highlights autonomous navigation and exploration in underground, GPS-denied environments. The challenge involves three types of environments: tunnel systems, urban underground, and cave networks. Teams deploy a fleet of autonomous vehicles to search for artifacts (backpack, cellphone, etc) and report their locations. The challenge allows a human operator to command the vehicles from the entrance of the competition site over intermittent wireless networks.


\begin{figure}[b]
	\vspace{-0.15in}
    \centering
    \includegraphics[width=0.8\linewidth]{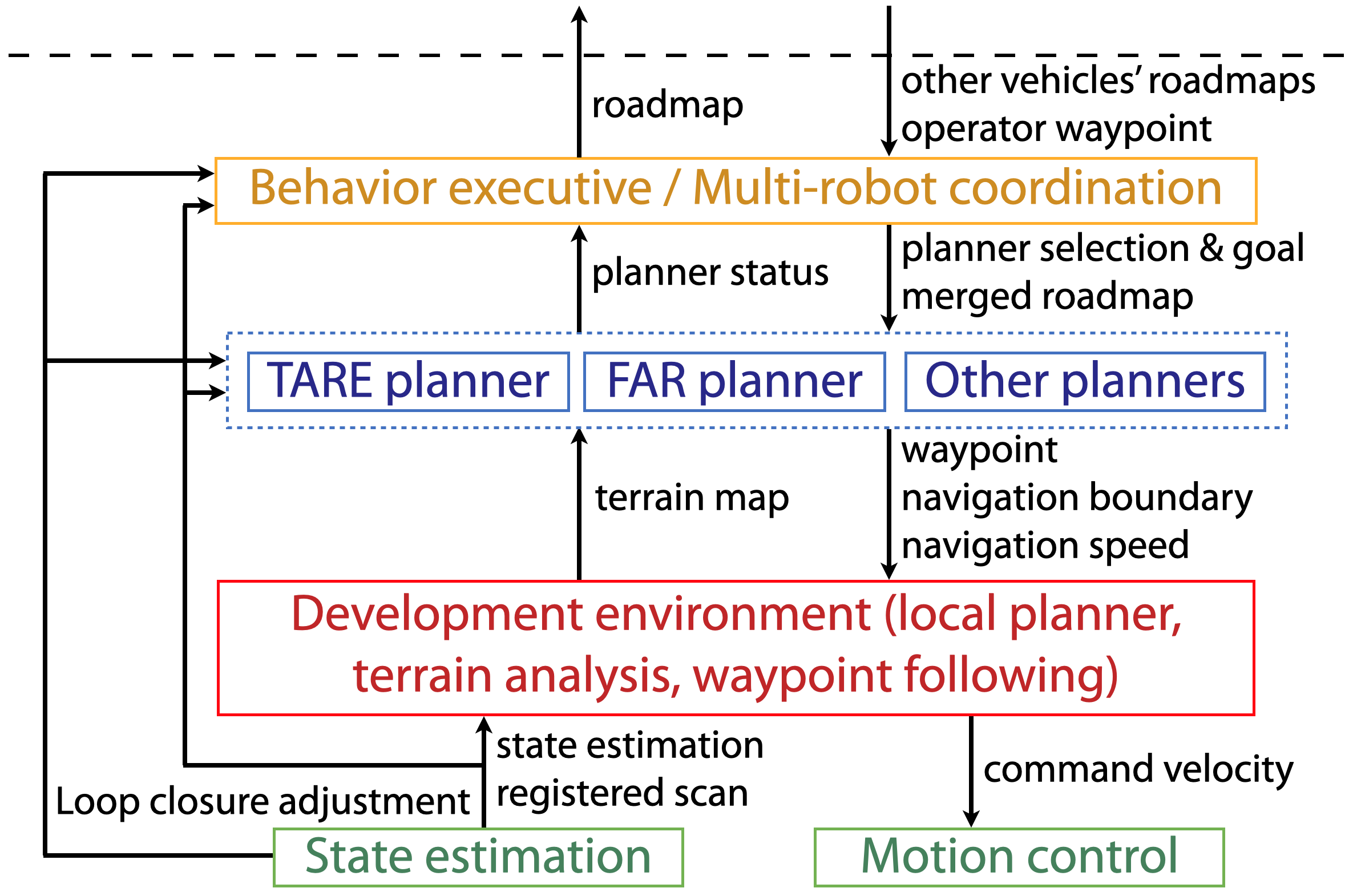}
    \vspace{-0.05in}
    \caption{System diagram for the ground vehicles used by the CMU-OSU Team in DARPA Subterranean Challenge. The system is extended from our basic system in Fig.~\ref{fig:system_diagram}.}
	\label{fig:subt_system}
\end{figure}

\begin{figure}[t]
	\vspace{0.05in}
    \centering
    \includegraphics[width=0.55\linewidth]{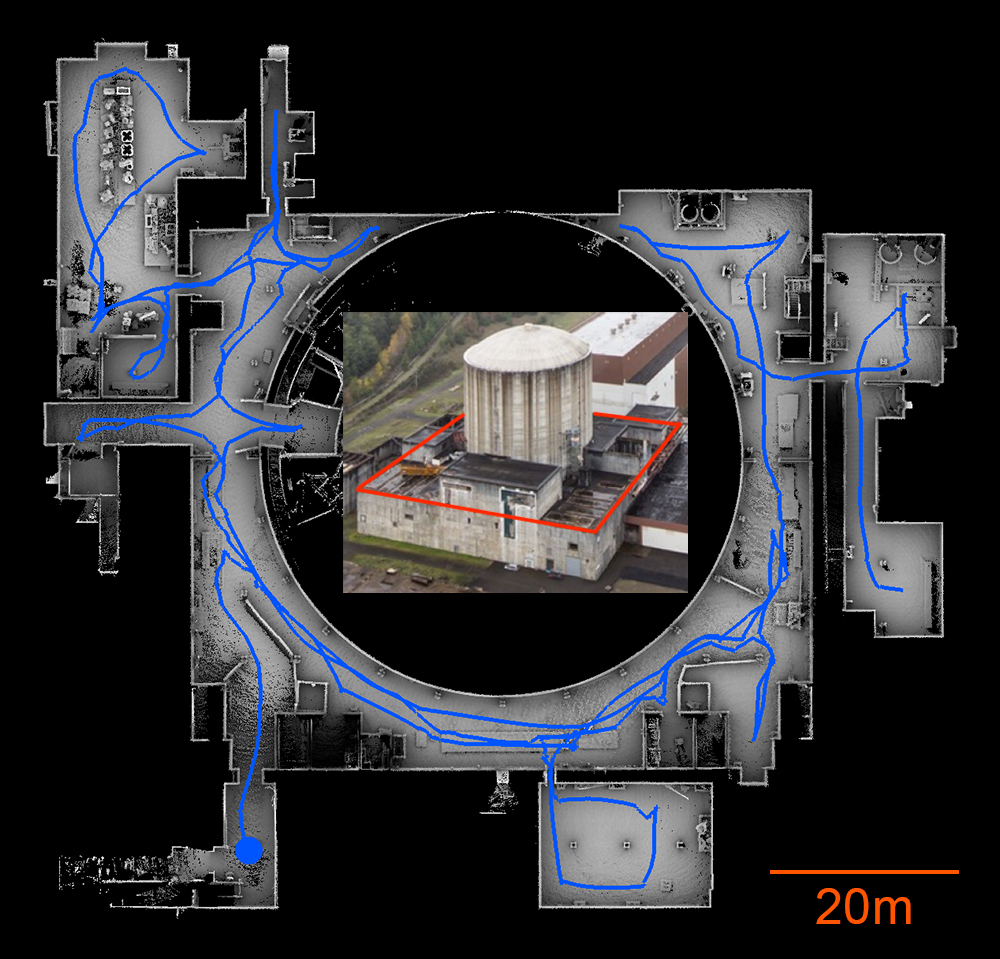}
	\vspace{-0.05in}
    \caption{Exploration result from DARPA Subterranean Challenge using TARE planner. The photo shows the exterior of the site where the event takes place (Urban Circuit at Satsop Nuclear Plant, WA). Our vehicle travels over 886m in 1458s to explore the entire floor.}
	\label{fig:satsop}
	\vspace{-0.15in}
\end{figure}

The ground vehicles used by the CMU-OSU Team are 4-wheel-drive and skid-steer platforms as shown in Fig.~\ref{fig:robots}(b). The navigation system is an extended version of our basic system. The system diagram is shown in Fig.~\ref{fig:subt_system}. The state estimation module can detect and introduce loop closures. The module outputs state estimation in the odometry frame generated by 3D Lidar-based odometry containing accumulated drift. When loop closure is triggered, it outputs loop closure adjustments to globally relax the vehicle trajectory and corresponding maps. Loop closure adjustments are used by the high-level planners since they are in charge of planning at the global scale. Modules such as local planner and terrain analysis only care about the local environment surrounding the vehicle and work in the odometry frame. The local planner and terrain analysis modules are extended to handle complex terrains including negative obstacles such as cliffs, pits, and water puddles with a downward-looking depth sensor. The TARE planner, FAR planner, and other planners (for stuck recovery, etc) are run in parallel for tasks such as exploration, go-to waypoint, and return home. On top of these planners, behavior executive and multi-robot coordination modules are built specifically for the challenge. The modules share explored and unexplored areas across multi-robots and call TARE and FAR planners cooperatively. In particular, when a long-distance transition is determined due to new areas containing artifacts are discovered or operator waypoints are received, the behavior executive switches to FAR planner for relocating the vehicle. During the relocation, FAR planner uses a sparse roadmap merged from multi-robots for high-level guidance. After the relocation is complete, TARE planner takes place for exploration.


\begin{figure}[t]
	\vspace{0.05in}
    \centering
    \includegraphics[width=0.85\linewidth]{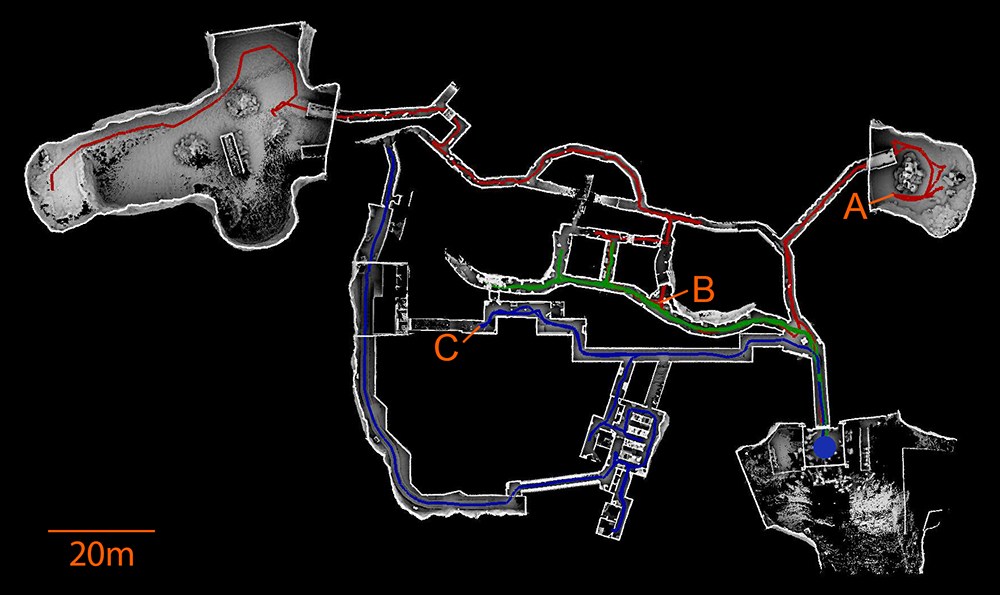}
	\vspace{-0.05in}
    \caption{Result from DARPA Subterranean Challenge Final Competition in Louisville Mega Cavern, KY. Three vehicles with red, green, and blue trajectories are deployed running TARE and FAR planners together. The segments between A and B on the red trajectory and between the start point (blue dot) and C on the blue trajectory use FAR planer to transit. Other segments on the trajectories use TARE planner to explore.}
	\label{fig:subt_final}
	\vspace{-0.15in}
\end{figure}

Fig.~\ref{fig:satsop} shows a representative exploration result from a competition run in Urban Circuit held at Satsop Nuclear Plant, WA. Our vehicle uses TARE planner to explore the environment fully autonomously, traveling over 886m in 1458s. 
Fig.~\ref{fig:subt_final} shows the result from the final competition in Louisville Mega Cavern, KY. The course combines tunnel, urban, and cave settings with complex topology. Our three vehicles use TARE and FAR planners together to traverse the environment. On the red trajectory, the vehicle first uses TARE planner to explore to B. Then, a request is received to transit to C using FAR planner. Upon arriving at C, the vehicle resumes exploration using TARE planner for the rest of the run. On the green trajectory, the vehicle uses TARE planner solely. On the blue trajectory, the vehicle transits to A using FAR planner and switches to TARE planner to explore thereafter. The three vehicles travel over 596.6m, 499.8m, and 445.2m respectively over a time span of 2259s.

\section{Extended Application Examples}\label{sec:ext_application}

Our system can contribute to the research society of semantic navigation where existing development environments are generally limited to datasets or simulation with relatively primitive navigation planning support. Several learning-based approaches can utilize the system for training purposes.


\textit{Self-supervised learning for visual navigation}: Learning-based navigation in the context of exploring a new environment, traversing to a goal point, or searching for a specific object can utilize our system to collect training data. For example, users can set up our Lidar-based system as a way of self-supervision for training a vision-based deep network, and then, during test time, the method uses only a camera to navigate. Since we support photorealistic environment models such as  Matterport3D, our simulation environment is designed to facilitate the simulation to real adaptation.


\textit{Mixed-initiative navigation with human instructions}: This can benefit from the Matterport3D house models which have semantic ground truth provided. By interpolating natural language, e.g., ``go through the door in the front and then turn left,'' users can use our system together with the semantic ground truth to locate the door, navigate the vehicle through the door, and then turn left. The algorithm being developed does not have access to the semantic ground truth but is trained by the reference trajectories from our system.

\textit{Social navigation in crowed environments}: We use Gazebo simulator which makes modifications to the simulator configuration convenient. Users can insert custom human avatars into the simulation environment to help develop social navigation algorithms. Once the human avatars are inserted, users have access to the human pose ground truth, which can potentially simplify the algorithm development by skipping the human pose detection module in simulation.

\section{Discussion and Future Activities}\label{sec:conclusion}

The purpose of this work is to provide a generic navigation system to the society to support a variety of multi-disciplinary research. To this end, we choose to work with ground vehicles instead of aerial vehicles due to their capacity to carry heavy sensor-computer payloads and extended battery life, since contemporary AI research often involves sophisticated sensors and computers with powerful GPUs. In the future, we plan to extend our system to multi-robot coordination and exploration. We also plan to organize tutorials on future ICRA and CVPR conferences to better prepare users to utilize our system in facilitating their work.

\bibliographystyle{IEEEtran}
\bibliography{references}

\end{document}